# Critical Remarks on Single Link Search in Learning Belief Networks

Y. Xiang        S.K.M Wong        N. Cercone
Department of Computer Science, University of Regina, Regina, Saskatchewan, Canada S4S 0A2

## Abstract

In learning belief networks, the single link lookahead search is widely adopted to reduce the search space. We show that there exists a class of probabilistic domain models which displays a special pattern of dependency. We analyze the behavior of several learning algorithms using different scoring metrics such as the entropy, conditional independence, minimal description length and Bayesian metrics. We demonstrate that single link lookahead search procedures (employed in these algorithms) cannot learn these models correctly. Thus, when the underlying domain model actually belongs to this class, the use of a single link search procedure will result in learning of an incorrect model. This may lead to inference errors when the model is used. Our analysis suggests that if the prior knowledge about a domain does not rule out the possible existence of these models, a multi-link lookahead search or other heuristics should be used for the learning process.

## 1 INTRODUCTION

As many effective probabilistic inference techniques have been developed [Pearl 86, Henrion 88, Lauritzen 88, Jensen 90, Xiang 93] and the applicability of probabilistic networks have been amply demonstrated in many artificial intelligence domains [Charniak 91], many researchers turn their attention to learning of such networks from data [Rebane 87, Herskovits 90, Buntine 91, Spirtes 91, Cooper 92, Lam 94, Heckerman 95].

It was shown that learning probabilistic networks is NP-hard [Bouckaert 94, Chickering 95]. Therefore, using heuristic methods in learning is justified. Many such algorithms adopt a scoring metric and a search procedure. The scoring metric is used to evaluate the goodness-of-fit of a structure to the observed data. The search procedure generates alternative structures and selects the best structure based on the evaluation.

Since the number of possible structures becomes very large as the number of links to add or delete in each search step increases, single link lookahead search is used in most learning algorithms to reduce the search space.

In this paper, we analyze the behavior of several algorithms in learning a class of probabilistic domain models (PDMs) called the pseudo-independent (PI) models. These algorithms use different scoring metrics such as the entropy, conditional independence, minimal description length, and Bayesian metrics. We show that single link lookahead search procedures cannot learn a PI model correctly. Thus, when the underlying PDM is actually a PI model, the use of such a procedure will result in learning of an incorrect model. This may lead to inference errors when the model is used. Therefore, multi-link search or other heuristics should be used if our prior knowledge about a domain does not eliminate the possible existence of a PI model.

The paper is organized as follows. In Section 2, we introduce terminologies used in the paper. The existence of PI models is shown in Section 3. In Section 4, we analyze the behavior of several algorithms for learning a Bayesian network (BN) as an approximate independence map (I-map) when the underlying PDM is a PI model. We summarize our results in Section 5.

## 2 TERMINOLOGY

Let $N$ be a set of discrete variables in a problem domain and $V \subseteq N$. A *configuration* $v$ of $V$ is an assignment of values to every variable in $V$. A *probabilistic domain model* (PDM) over $N$ is an encoding of probabilistic information that defines the probability of every configuration of $V$ for every $V \subseteq N$. A PDM over $N$ can be specified by a joint probability distribution (jpd) over $N$ denoted by $P(N)$. We denote the marginal distribution over $V \subseteq N$ by $P(V)$. The entropy of $V$ is defined by $H(V) = -\sum_v P(v) \log(P(v))$.

Let $U, V, Z \subset N$ be disjoint subsets. $U$ and $V$ are *conditionally independent* given $Z$, denoted $Ind(U, Z, V)$, iff $P(u|v\ z) = P(u|z)$ whenever $P(v\ z) > 0$.

Since we use graphs to represent independence relations among variables, we shall use *nodes* and *variable*

interchangeably. Consider three disjoint subsets $U$, $V$ and $Z$ of nodes in a graph. We use $<U|Z|V>$ to denote that $Z$ *separates* $U$ from $V$ in a graph (directed or undirected). In an undirected graph, this means that nodes in $Z$ intercept all the paths between nodes of $U$ and nodes of $V$. In a directed graph, this means that $Z$ *d-separates* $U$ from $V$ [Pearl 88]. We shall refer to an edge in any graph as a *link*. When the direction of a link in a graph is of concern, we refer to it as an *arc*.

A graph $G$ is an *independence map (I-map)* of a PDM $\mathcal{M}$ over $N$ if there is an one-to-one correspondence between nodes of $G$ and variables in $N$ such that for all disjoint subsets $U$, $V$ and $Z$ of $N$, $<U|Z|V> \Rightarrow Ind(U, Z, V)$. That is, in an I-map, variables that are graphically separated are independent. Variables not graphically separated, however, are not necessarily dependent. A *minimal* I-map is an I-map in which no link can be deleted such that it is still an I-map. See [Pearl 88] for more details on graphical representation of dependency models.

A BN $(G, P(N))$ over a set $N$ of variables in a problem domain consists of an I-map $G$ of the domain and a distribution $P(N)$. $G$ is a directed acyclic graph (DAG) whose nodes are labeled by the variables in $N$. For each node $X \in N$, we denote its parent set by $\Pi_X$. $P(N)$ is a jpd specified by conditional probabilities of each variable $X \in N$ conditioned on $\Pi_X$.

## 3  PI MODELS

We show that there exists a class of PDMs which display a special pattern of dependence relations.

**Theorem 1** *For any integer $\eta \geq 3$, there exists a collection $C$ of probabilistic domain models over a set $N$ of $\eta$ binary variables such that the following hold for each $\mathcal{M} \in C$.*

**S1** *For each $Y \in N$, $P_\mathcal{M}(N \backslash \{Y\}) = \prod_{X \in N, X \neq Y} P_\mathcal{M}(X)$.*

**S2** *For each pair $X, Y \in N$ and $X \neq Y$, $Ind(\{X\}, N \backslash \{X, Y\}, \{Y\})$ does not hold in $\mathcal{M}$.*

*We shall refer to each $\mathcal{M}$ as a* **pseudo-independent** *(PI) model.*

Before proving the theorem, we intuitively describe the dependency pattern displayed by the PI models. S2 implies that no pair of variables of $N$ are independent given all other variables. Therefore, in any I-map $G_\mathcal{M}$ of $\mathcal{M}$, there must be a *direct* line between each pair of them, i.e., $G_\mathcal{M}$ is a complete graph. We say that variables in such PDMs are *collectively dependent*. On the other hand, S1 implies that variables in any subset of $N$ of size $\eta - 1$ are *pairwise marginally independent*.

Proof:

It is sufficient to construct a parameterized jpd given $\eta$ such that both S1 and S2 hold and the parameter can take on infinite possible values.

Let $X_1, \ldots, X_\eta$ denote $\eta$ binary variables in $N$ and $P(X_{i,0}) = P(X_{i,1}) = 0.5$ $(i = 1, \ldots, \eta)$ where $X_{i,0}$ and $X_{i,1}$ are the two outcomes of $X_i$. There are exactly $\eta$ distinct subsets of $N$ of size $\eta - 1$. For each subset $\{X_{i_1}, \ldots, X_{i_{\eta-1}}\}$ where $1 \leq i_j \leq \eta$, S1 is equivalent to

$$P(X_{i_1}, \ldots, X_{i_{\eta-1}}) = 0.5^{\eta-1}.$$

We have omitted the second index because the particular configuration does not affect the probability value. Models that satisfy S1 do exist. A jpd $P^*(N) = P(X_1, \ldots, X_\eta) = 0.5^\eta$ is one example. However, $P^*(N)$ does not satisfy S2. We will construct a jpd of $\mathcal{M}$ which satisfies both S1 and S2.

We can view the above condition, which is equivalent to S1, as a constraint

$$P(X_{i_1}, \ldots, X_{i_{\eta-1}}) = P(X_{i_1}, \ldots, X_{i_{\eta-1}}, X_{i_\eta, 0}) +$$
$$P(X_{i_1}, \ldots, X_{i_{\eta-1}}, X_{i_\eta, 1}) = 0.5^{\eta-1}$$

on the subset $\{X_{i_1}, \ldots, X_{i_{\eta-1}}\}$. We therefore have $\eta$ constraints, one for each such subset.

To construct a desired jpd, we assign a probability value to each of the $2^\eta$ configurations, each of which is denoted by a binary $\eta$-tuple. For example, the configuration $(X_{1,0}, \ldots, X_{\eta,0})$ is denoted $(0, \ldots, 0)$. We group the tuples according to the number of 1s contained in each tuple and index the groups as $GP_0, \ldots, GP_\eta$. For example, $GP_0$ has a single tuple $(0, \ldots, 0)$, $GP_1$ has $\eta$ tuples $(0, \ldots, 0, 1)$, $(0, \ldots, 0, 1, 0)$, ..., and $(1, 0, \ldots, 0)$.

We assign probability values to the configurations group by group in ascending order of the group index. To make a new assignment, we check the configurations whose probability values have been assigned, determine how many of the $\eta$ constraints are involved in the assignment, and ensure that the new assignment conform to the constraints.

We start by assigning the single configuration in $GP_0$: $P(0, \ldots, 0) = 0.5^{\eta-1}q$, where $q \in [0, 1]$ and $q \neq 0.5$. This assignment does not violate any constraints. We then assign a configuration in $GP_1$:

$$P(0, \ldots, 0, 1) = P(X_{1,0}, \ldots, X_{\eta-1,0}) -$$
$$P(X_{1,0}, \ldots, X_{\eta-1,0}, X_{\eta,0}) = 0.5^{\eta-1}(1-q).$$

Note that this assignment involves only one constraint and involves the only configuration whose value has been assigned. We will say that the assignment of probability value to configuration $(0, \ldots, 0, 1)$ involves the constraint *relative to* the configuration $(0, \ldots, 0, 0)$.

We make the following observation: If $c_1$ is a configuration whose probability has been assigned and $c_2$ is a configuration whose probability is to be assigned, then the assignment involves a constraint relative to $c_1$ if and only if $c_1$ and $c_2$ differ by exactly one attribute.

This observation leads to two implications. First, the assignment of $c_2$ cannot involve a constraint relative to another configuration in the same group, since configurations in the same group differ by at least two attributes. For example, $(0, \ldots, 0, 1)$ and $(0, \ldots, 1, 0)$ in $GP_1$, and $(0, \ldots, 0, 1, 1)$ and $(0, \ldots, 1, 0, 1)$ in $GP_2$.

Second, if $c_2 \in GP_i$, the assignment of $c_2$ can only involve a constraint relative to configurations in $GP_{i-1}$. This is because configurations in $GP_j$ $(j \leq i-2)$ differ from $c_2$ by at least two 1s. Therefore, when we assign a configuration, we have only to check configurations in the very last group assigned. Note that the assignment may still involve multiple constraints each relative to a distinct configuration. For example, the assignment of $(0, \ldots, 0, 1, 1, 1)$ in $GP_3$ involves three constraints relative to $(0, \ldots, 0, 1, 1)$, $(0, \ldots, 0, 1, 1, 0)$ and $(0, \ldots, 0, 1, 0, 1)$ in $GP_2$, respectively. We show that all of the constraints involved can be satisfied simultaneously.

Each configuration in $GP_1$ involves a single constraint relative to the single configuration $(0, \ldots, 0)$ in $GP_0$. To satisfy each constraint, we assign the configuration $0.5^{\eta-1}(1-q)$ as we did in the second assignment above. Hence all configurations in $GP_1$ have the *same* probability value, since all distributions of $\eta - 1$ order have the same value $0.5^{\eta-1}$. Therefore, for each configuration $c \in GP_2$, even though it involves two constraints, each relative to a different configuration in $GP_1$, the assignment $P(c) = 0.5^{\eta-1}q$ satisfies both simultaneously.

Thus, by following this procedure, we can construct a jpd for $\mathcal{M}$ by alternating the assignment of $0.5^{\eta-1}q$ and $0.5^{\eta-1}(1-q)$ to configurations in successive groups. The resultant jpd clearly satisfies S1.

To show that the jpd also satisfies S2, we need to show, for an arbitrary pair $X_i, X_j$ $(i \neq j)$ and $W = N \setminus \{X_i, X_j\}$, that $P(X_i|X_j, W) \neq P(X_i|W)$, or equivalently, $P(X_i, X_j, W) \neq P(X_i|W)P(X_j, W)$. Since $P(X_i|W) = 0.5$ and $P(X_j, W) = 0.5^{\eta-1}$ by S1, we have $P(X_i|W)P(X_j, W) = 0.5^{\eta}$. However, $P(X_i, X_j, W)$ has the value $0.5^{\eta-1}q$ or $0.5^{\eta-1}(1-q)$, where $q \neq 0.5$.

We have now constructed a jpd that satisfies both S1 and S2, and has a parameter $q$. Since $q$ can take any value in the intervals $[0, 0.5)$ and $(0.5, 1]$, the theorem is proven.   □

Consider the following example of a PI model. Suppose we have a digital gate with three inputs $X_i$ $(i = 1, 2, 3)$ and an output $X_4$. The output $X_4 = 1$ whenever any two inputs are 0 and a third input is 1, or all inputs are 1. Suppose the three inputs are independent to each other and each of them has equal chance to be 0 or 1. Table 1 shows the jpd of these four variables. It can be easily verified that (1) the marginal distribution of each variable is 0.5, (2) any subset of two or three variables are mutually independent, and (3) the jpd is not $0.5^4 = 0.0625$.

Table 1: An example of a pseudo-independent model.

| $(X_1, X_2, X_3, X_4)$ | $P(.)$ | $(X_1, X_2, X_3, X_4)$ | $P(.)$ |
|---|---|---|---|
| (0,0,0,0) | 0.125 | (1,0,0,0) | 0 |
| (0,0,0,1) | 0 | (1,0,0,1) | 0.125 |
| (0,0,1,0) | 0 | (1,0,1,0) | 0.125 |
| (0,0,1,1) | 0.125 | (1,0,1,1) | 0 |
| (0,1,0,0) | 0 | (1,1,0,0) | 0.125 |
| (0,1,0,1) | 0.125 | (1,1,0,1) | 0 |
| (0,1,1,0) | 0.125 | (1,1,1,0) | 0 |
| (0,1,1,1) | 0 | (1,1,1,1) | 0.125 |

In the PI models constructed in the proof of Theorem 1, the marginal of each variable is equal to 0.5. However, PI models are not restricted to 0.5 marginals. Table 2 provides a jpd of three variables that have different marginals, in which (1) the marginals are $P(X_{1,0}) = 0.6$, $P(X_{2,0}) = 0.4$ and $P(X_{3,0}) = 0.2$, (2) any two variables are marginally independent, and (3) the jpd is not equal to the product $P(X_1)P(X_2)P(X_3)$.

Table 2: A PI model with different variable marginals.

| $(X_1, X_2, X_3)$ | $P(.)$ | $(X_1, X_2, X_3)$ | $P(.)$ |
|---|---|---|---|
| (0,0,0) | 0.024 | (1,0,0) | 0.056 |
| (0,0,1) | 0.216 | (1,0,1) | 0.104 |
| (0,1,0) | 0.096 | (1,1,0) | 0.024 |
| (0,1,1) | 0.264 | (1,1,1) | 0.216 |

Among all the PDMs, pseudo-independent models represent one extreme. The other extreme is represented by models which display a totally different pattern of dependence relations. In the I-map of those models, no pair of variables connected by a link displays marginal independence. Between these two extremes, a whole spectrum of pseudo-independent models exist, in which variables are collectively dependent, marginally independent in some pairs and not marginally independent in other pairs. To classify these models, we shall refer to the models in Theorem 1 as *full* PI models and the models between the two extremes as *partial* PI models. Table 3 depicts such a partial PI model of three variables. The marginal for each variable is 0.5. Any pair of variables are dependent given the third. However, $X_1$ and $X_2$ are marginally independent $(P(X_1, X_2) = P(X_1)P(X_2))$, so are $X_1$ and $X_3$, but $X_2$ and $X_3$ are *not* marginally independent, namely, $P(X_2, X_3) \neq P(X_2)P(X_3)$.

The PI models presented thus far are defined based on the entire domain of variables. In general, a PI model can be *embedded* as a submodel. Table 4 shows a PDM with four variables $X_i$ ($i = 1, 2, 3, 4$). It contains an embedded submodel identical to the partial PI model (of $X_1$, $X_2$ and $X_3$) given in Table 3. The marginal for each variable is 0.5 except $P(X_4 = 0) = 0.365$.

Table 3: A partial PI model.

| $(X_1, X_2, X_3)$ | $P(.)$ | $(X_1, X_2, X_3)$ | $P(.)$ |
|---|---|---|---|
| (0, 0, 0) | 0.225 | (1, 0, 0) | 0.20 |
| (0, 0, 1) | 0.025 | (1, 0, 1) | 0.05 |
| (0, 1, 0) | 0.025 | (1, 1, 0) | 0.05 |
| (0, 1, 1) | 0.225 | (1, 1, 1) | 0.20 |

The marginal for the subset $\{X_1, X_2, X_3\}$ is identical to that of Table 3, so the dependency relations among the three variables remain the same. But for variables $X_2$, $X_3$ and $X_4$, they are both collectively and pairwise dependent. The undirected minimal I-map of the PDM has each pair of variables connected except $X_1$ and $X_4$.

Table 4: An embedded PI model.

| $(X_1, X_2, X_3, X_4)$ | $P(.)$ | $(X_1, X_2, X_3, X_4)$ | $P(.)$ |
|---|---|---|---|
| (0, 0, 0, 0) | 0.0225 | (1, 0, 0, 0) | 0.02 |
| (0, 0, 0, 1) | 0.2025 | (1, 0, 0, 1) | 0.18 |
| (0, 0, 1, 0) | 0.005 | (1, 0, 1, 0) | 0.01 |
| (0, 0, 1, 1) | 0.02 | (1, 0, 1, 1) | 0.04 |
| (0, 1, 0, 0) | 0.0175 | (1, 1, 0, 0) | 0.035 |
| (0, 1, 0, 1) | 0.0075 | (1, 1, 0, 1) | 0.015 |
| (0, 1, 1, 0) | 0.135 | (1, 1, 1, 0) | 0.12 |
| (0, 1, 1, 1) | 0.09 | (1, 1, 1, 1) | 0.08 |

## 4 BEHAVIOR OF COMMON ALGORITHMS

In this section, we analyze the behavior of four algorithms for learning an approximate I-map of a PDM expressed as a BN. These algorithms use different scoring metrics: entropy, conditional independence, the minimal description length and the Bayesian, but they all employ single link lookahead search procedures. We show that when the data-generating PDM is a PI model, none of the algorithms can learn an I-map of the PDM correctly.

### 4.1 THE KUTATO ALGORITHM

The Kutato algorithm [Herskovits 90] developed by Herskovits and Cooper learns a BN using an entropy scoring metric and single link lookahead search. Given a BN $(G, P(N))$, the entropy of $N$ defined by $P(N)$ is

$$H(N) = \sum_{X \in N} (\sum_{\pi_X} P(\Pi_X = \pi_X) \, H(X|\pi_X)) \qquad (1)$$

where $\pi_X$ is a configuration of $\Pi_X$. $H(X|\pi_X)$ is defined below where $x$ is a value of $X$,

$$H(X|\pi_X) = -\sum_x P(X = x|\pi_X) \, lnP(X = x|\pi_X).$$

Kutato starts with an empty graph. At each step of the search, the algorithm evaluates $H(N)$ for each possible arc that may be added subject to a given ordering of variables, and selects the arc resulting in the minimum value of $H(N)$ to add to the current BN.

To simplify our presentation, suppose the data-generating PDM is a full PI model, i.e., variables are pairwise marginally independent and collectively dependent. The case for an embedded PI submodel can be similarly shown. Consider an arbitrary variable $X$. Suppose at a particular step of the search, no parent node has been connected to $X$ ($\Pi_X = \phi$). Then the entropy contribution of $X$ (the term associated with $X$ in Equation (1)), which we shall denote by $H_X$, is

$$H_X = H(X) = -\sum_x P(X = x) \, lnP(X = x).$$

Now suppose a potential parent variable $Y$ is considered by Kutato. After adding the arc $(Y, X)$ to the BN, the new entropy contribution of $X$ is

$$H'_X = \sum_y P(Y = y) \, H(X|y).$$

Since $X$ and $Y$ are marginally independent, we obtain

$$H(X|y) = -\sum_x P(X = x|y) \, lnP(X = x|y)$$

$$= -\sum_x P(X = x) \, lnP(X = x) = H(X),$$

which yields $H'_X = \sum_y P(Y = y) \, H(X) = H(X) \sum_y P(Y = y) = H(X) = H_X$. Because Kutato uses the single-step lookahead search, the entropy contributions of other variables remain the same and therefore $H(N)$ remains the same after the arc $(Y, X)$ is added. Since the arc $(Y, X)$ does not decrease $H(N)$, it will be rejected by Kutato.

Since this argument applies to any $X \in N$ and any $Y \in N$ such that $Y > X$ in the ordering, we conclude that Kutato will return a BN without connecting any pair of variables. That is, it fails to learn the collective dependency of the PI model.

### 4.2 THE LAM-BACCHUS ALGORITHM

Likewise, PI models cannot be learned by the algorithm suggested by Lam and Bacchus [Lam 94], which learns a BN based on the minimal description length (MDL) principle.

The algorithm first computes the *average mutual information* $I(X; Y)$ between each pair of nodes $X$ and $Y$ (corresponding to an undirected link with the direction to be determined later):

$$I(X; Y) = \sum_{x,y} P(X = x, Y = y) \, ln \frac{P(X = x, Y = y)}{P(X = x)P(Y = y)}.$$

The algorithm then places all links in a list in descending order of the mutual information between the end nodes. The candidate BNs are generated by starting with an empty graph and including one link at a time from the beginning of the link list and traversing down the list. It allocates equal amount of computational resources to explore candidate BNs of identical number of links (having the same complexity). After each complexity class has exhausted its resources, the best candidate BN, according to the Kullback-Leibler cross entropy [Kullback 51] scoring metric, is chosen. The BN that has the minimal description length across classes will be the final output.

Suppose the data-generating model has a partial PI submodel embedded. It is well known that $I(X;Y) = 0$ if and only if $X$ and $Y$ are independent. If a pair of nodes in the submodel are marginally independent, then the link between them has zero mutual information. These links will be placed at the end of the link list and will be the last to be included in any candidate BNs. If these BNs are ever considered, the algorithm must have exhausted almost all possible BNs, which has an exponential complexity. Therefore, in practice, these BNs would have no chance to be tested and selected as the final output.

### 4.3 THE PC ALGORITHM

The previous two algorithms start with an empty graph. In contrast, the algorithm PC [Spirtes 91] developed by Spirtes and Glymour learns a BN by starting from a complete graph.

In the first pass, the algorithm removes each link if the end nodes of the link are *marginally* independent. In the second pass, it removes each link if the end nodes of the link are independent conditioned on a third node. In each of the following passes, it removes each link if the end points of the link are independent conditioned on a subset of nodes of higher order until a stopping condition is met.

If the data-generating PDM has a partial PI submodel embedded, then some pair of nodes in the submodel are marginally independent. The links between each pair of them will be deleted in the first pass of the search. Therefore the collective dependency of the submodel will not be reflected in the final learned BN.

### 4.4 THE K2 ALGORITHM

The K2 algorithm [Cooper 92] suggested by Cooper and Herskovits learns the structure of a BN based on a Bayesian score. That is, it tries to learn structures that have the highest probability given a database $D$ of cases.

To evaluate dependency between a variable $X \in N$ and a potential parent set $\Pi_X$, K2 uses the following scoring function

$$g(X, \Pi_X) = \prod_{\pi_X} (\frac{(|X|-1)!}{(N_{\pi_X} + |X| - 1)!} \prod_x N_{x,\pi_X}!) \quad (2)$$

where $|X|$ is the number of values that $X$ takes in $D$, $x$ is one such value, $\pi_X$ is a configuration that $\Pi_X$ takes in $D$, $N_{\pi_X}$ is the number of cases in $D$ in which $\Pi_X = \pi_X$, and $N_{x,\pi_X}$ is the number of cases in $D$ in which $X = x$ and $\Pi_X = \pi_X$. Let $\Pi_X$ be the parent set of $X$ in the current structure during the search. To search for the next parent variable $Y$ of $X$, K2 evaluates $g(X, \Pi_X \cup \{Y\})$ for each variable $Y$ such that $Y > X$ in a given ordering. The variable $Y$ that maximizes $g(X, \Pi_X \cup \{Y\})$ such that $g(X, \Pi_X \cup \{Y\}) > g(X, \Pi_X)$ will be chosen as a new parent of $X$.

We do not realize any reference with a precise interpretation of $g(X, \Pi_X)$. Hence, it is not obvious how the value of $g(X, \Pi_X)$ changes when $X$ and $\Pi_X$ are independent and when they are not. Consequently, the behavior of K2 in learning a PI model cannot be derived in a straightforward fashion as in the previous algorithms. To establish this behavior of K2, we first show that K2 behaves the same no matter whether $X$ and $\Pi_X$ are truly independent or pseudo-independent. To eliminate the effect of sampling, for now, we make the following assumption which will be relaxed later.

**Assumption 1** *For each variable $X$ and its parent set $\Pi_X$, the database $D$ satisfies $N_{X,\Pi_X} = m\ P(X, \Pi_X)$, where $m$ is the total number of cases in $D$.*

Using Assumption 1, we obtain

$$g(X, \Pi_X) = \prod_{\pi_X} (\frac{(|X|-1)!}{(m\ P(\Pi_X = \pi_X) + |X| - 1)!} \prod_x (m\ P(X = x, \Pi_X = \pi_X))!). \quad (3)$$

As K2 uses the single link lookahead search and starts with an empty graph, let us consider an arbitrary variable $X$ currently with $\Pi_X = \phi$. Suppose a potential parent variable $Y$ is to be evaluated next. Before $Y$ is added to the empty $\Pi_X$, we have

$$g(X, \phi) = \frac{(|X|-1)!}{(m + |X| - 1)!} \prod_x (m\ P(X = x))!. \quad (4)$$

After $Y$ is added to $\Pi_X$,

$$g(X, \{Y\}) = \prod_y (\frac{(|X|-1)!}{(m\ P(Y = y) + |X| - 1)!}$$

$$\prod_x (m\ P(X=x, Y=y))!). \quad (5)$$

Suppose a variable $Y^*$ has the highest score among variables larger than $X$ in the node ordering, i.e.,

$$g(X, \{Y^*\}) = \max_{Y>X} g(X, \{Y\}).$$

Then K2 will choose $Y^*$ as a new parent of $X$ if and only if $g(X, \{Y^*\}) > g(X, \phi)$. From Equations (4) and (5), clearly this decision depends entirely on the marginal distribution $P(X, Y^*)$. As mentioned in Section 3, $P(X, Y^*) = P(X)\ P(Y^*)$ may hold either because $X$ and $Y^*$ are truly independent or because they are pseudo-independent. Therefore, K2 will behave the same way in both cases. That is, K2 will either reject $Y^*$ as the parent of $X$ in both cases, which is incorrect in the case of pseudo-independence, or accept $Y^*$ as the parent of $X$ in both cases, which is incorrect in the case of true independence. However, at this moment, it is still unclear which mistake K2 will make.

Next, we show that K2 will actually reject $Y^*$ as the parent in both cases. To simplify our presentation, we assume that $X$ and $Y^*$ are binary variables and we denote their values by 0 and 1.

**Assumption 2** *Variables $X$ and $Y^*$ are binary.*

Applying Assumption 2, we simplify the notation:

$$\begin{array}{ll}
w = m\ P(X=0) & m-w = m\ P(X=1) \\
v = m\ P(Y^*=0) & m-v = m\ P(Y^*=1) \\
u = m\ P(X=0, Y^*=0) & v-u = m\ P(X=1, Y^*=0) \\
z = m\ P(X=0, Y^*=1) & m-v-z = m\ P(X=1, Y^*=1).
\end{array} \quad (6)$$

Note that $w$, $v$, $u$ and $z$ are numbers of cases in $D$ and thus are integers. Using Equation (6), we can rewrite $g(X, \phi)$ and $g(X, \{Y^*\})$ as follows.

$$g(X, \phi) = \frac{w!\ (m-w)!}{(m+1)!} \quad (7)$$

$$g(X, \{Y^*\}) = \frac{u!\ (v-u)!\ z!\ (m-v-z)!}{(v+1)!\ (m-v+1)!} \quad (8)$$

For any real number $r$, we shall use $\lfloor r \rfloor$ to denote the greatest integer less than or equal to $r$. Due to the symmetry between $w$ and $m-w$, $v$ and $m-v$, $u$ and $v-u$, and $z$ and $m-v-z$, we may assume

$$w \leq \lfloor m/2 \rfloor \quad v \leq \lfloor m/2 \rfloor \quad u \leq \lfloor v/2 \rfloor \quad z \leq \lfloor (m-v)/2 \rfloor. \quad (9)$$

Since we are interested in the cases when $X$ and $Y^*$ display probabilistic independence, we do not consider the cases when either one of them is deterministic, which is expressed by the following assumption.

**Assumption 3** $P(X)$ and $P(Y^*)$ are strictly positive.

From this assumption, we have $u > 0$ and $v - u > 0$ and hence $v \geq 2$. Similarly, from $u > 0$ and $z > 0$, we derive $w \geq 2$ and hence $m \geq 4$. We summarize these in the following inequalities.

$$w \geq 2 \qquad v \geq 2 \qquad m \geq 4. \quad (10)$$

To study the behavior of K2 when $X$ and $Y^*$ are (truly or pseudo) independent, we assume the following.

**Assumption 4** $P(X, Y^*) = P(X)\ P(Y^*)$ *holds.*

The above assumption implies the following relation

$$\frac{w}{m-w} = \frac{u}{v-u} = \frac{z}{m-v-z}. \quad (11)$$

Using this relation, we can eliminate $u$ and $z$ from Equation (8) to obtain

$$g(X, \{Y^*\}) = \frac{(vw/m)!\ (v(m-w)/m)!}{(v+1)!}$$

$$\frac{((m-v)w/m)!\ ((m-v)(m-w)/m)!}{(m-v+1)!}.$$

We define the ratio $r = g(X, \phi)/g(X, \{Y^*\})$ and expand it as follows:

$$r = \frac{w!\ (m-w)!\ (v+1)\ v!}{(m+1)\ m!\ (vw/m)!\ (v(m-w)/m)!}$$

$$\frac{(m-v+1)\ (m-v)!}{((m-v)w/m)!\ ((m-v)(m-w)/m)!} \quad (12)$$

We show that under Assumptions 1 through 4, it is the case that $r > 1$. First, we derive a lower bound of $r$ suitable for analysis in real by applying the well known Stirling's formula [Wang 79]

$$(2\pi)^{0.5}\ \frac{n^{n+0.5}}{e^n} < n! < (2\pi)^{0.5}\ \frac{n^{n+0.5}}{e^n}\ \frac{1}{1-\frac{1}{12n}}.$$

Using this formula, we substitute each factorial in the numerator in Equation (12) by its lower bound and each factorial in the denominator by its upper bound. We obtain

$$r > \frac{(v+1)\ (m-v+1)\ m^{1.5}}{(2\pi)^{0.5}\ v^{0.5}\ (m-v)^{0.5}\ w^{0.5}\ (m-w)^{0.5}\ (m+1)}$$

$$\times (1 - \frac{1}{12m})(1 - \frac{m}{12vw})(1 - \frac{m}{12v(m-w)})$$

$$\times (1 - \frac{m}{12(m-v)w})(1 - \frac{m}{12(m-v)(m-w)}).$$

By replacing $v$ and $m-v$ in the denominator with $v+1$ and $m-v+1$, respectively, we can derive a simpler lower bound of $r$ denoted by $r'$, namely:

$$r > \frac{(v+1)^{0.5} (m-v+1)^{0.5} m^{1.5}}{(2\pi)^{0.5} w^{0.5} (m-w)^{0.5} (m+1)} (1 - \frac{1}{12m})$$
$$\times (1 - \frac{m}{12vw})(1 - \frac{m}{12v(m-w)})(1 - \frac{m}{12(m-v)w})$$
$$\times (1 - \frac{m}{12(m-v)(m-w)}) \equiv r'. \quad (13)$$

To find the lower bound of $r'$ for all values of $m$, $w$ and $v$, we first analyze the lower bound of $r'$ given $m$ and $w$. We show that $r'$ is a monotonic increasing function of $v$ given $m$ and $w$, and therefore it reaches its minimum value when $v = 2$. Treating $m$ and $w$ as constants, we extract the factors of $r'$ that contains $v$ to construct a new function $h(v)$, defined by:

$$h(v) \equiv (v+1)^{0.5} (m-v+1)^{0.5} \frac{12vw - m}{v} \frac{12v(m-w) - m}{v}$$
$$\times \frac{12(m-v)w - m}{m-v} \frac{12(m-v)(m-w) - m}{m-v}. \quad (14)$$

To show that $r'$ is monotonic increasing with $v$, it is sufficient to show that $h(v)$ is monotonic increasing. This is shown in the following lemma.

**Lemma 2** *The function $h(v)$ defined by Equation (14) is monotonic increasing in $[2, m/2]$, where $m \geq 4$ and $w \in [2, m/2]$.*

Proof:

Let $h(v) = h_1(v) \, h_2(v) \, h_3(v)$, where

$$h_1(v) \equiv (v+1)^{0.5} (m-v+1)^{0.5}$$
$$h_2(v) \equiv \frac{12vw - m}{v} \frac{12(m-v)w - m}{m-v}$$
$$h_3(v) \equiv \frac{12v(m-w) - m}{v} \frac{12(m-v)(m-w) - m}{m-v}.$$

To show $h(v)$ is monotonic increasing, it suffices to show that $h_i(v)$ ($i = 1, 2, 3$) are all monotonic increasing. The first order derivatives of $h_i(v)$ are:

$$h'_1(v) = \frac{0.5 (m - 2v)}{(v+1)^{0.5} (m-v+1)^{0.5}},$$
$$h'_2(v) = \frac{m^2 (m - 2v) (12w - 1)}{v^2 (m-v)^2},$$
$$h'_3(v) = \frac{m^2 (m - 2v) (12(m-w) - 1)}{v^2 (m-v)^2}.$$

Clearly, all these derivatives are positive in $[2, m/2)$ and thus the lemma is proven.   □

Lemma 2 implies the following corollary.

**Corollary 3** *Given $m$ and $w$ where $m \geq 4$ and $w \in [2, m/2]$, the function $r'$ as defined by Equation (13) is monotonic increasing with $v \in [2, m/2)$ and reaches its minimum value when $v = 2$.*

Note that Corollary 3 is true for any real values of $m$ and $w$ as long as they are within the ranges stated in the corollary. We now restrict the values of $w$ to those compatible with conditions we have established. From Assumption 3 and $v = 2$, we derive $u = v - u = 1$. From Equation (11), we obtain $w = m - w$ which implies $w = m/2$. Therefore, $r'$ reaches its minimum when $v = 2$ and $w = m/2$. Using this result, we substitute $w$ by $m/2$ and $v$ by 2 in $r'$ to obtain

$$\min(r') = \frac{121}{24 (m+1)} \left(\frac{m(m-1)}{6\pi}\right)^{0.5} (1 - \frac{1}{12m}) (1 - \frac{1}{6(m-2)})^2.$$

Clearly, $\min(r')$ is monotonically increasing with $m$. Since $\min(r') = 1.0096$ when $m = 14$, we conclude that $r > r' > 1$ for all $m \geq 14$.

When $m < 14$, $\min(r')$ may be smaller than 1. For example, $\min(r') = 0.9855$ when $m = 12$ even though $\min(r) = 1.3846$ since $\min(r')$ is a lower bound of $\min(r)$. Through exhaustive testing, we found that for all $m$ ($4 \leq m < 14$) such Assumption 1 through 4 hold, it is the case that $\min(r) > 1$. We summarize the result as follows.

**Proposition 4** *If Assumption 1 through 4 hold, then $g(X, \phi) > g(X, \{Y^*\})$.*

Proposition 4 is derived with the help of Assumption 1. In practice, the assumption does not hold due to sampling. However, if the number $m$ of cases in $D$ is sufficiently large, $N_{X, \Pi_X}$ will be close to $m P(X, \Pi_X)$. Our analysis will still apply if we substitute $v$ by $v' + \Delta v$ where $v' = N_{Y^* = 0}$ and $\Delta v$ is an error term, and do the same with other parameters. Since $r'$ is a continuous function, we will still have $r > 1$ when $\Delta$ error terms are small. We may thus drop Assumption 1:

**Proposition 5** *If for each variable $X$ and its parent set $\Pi_X$, the database $D$ satisfies $N_{X, \Pi_X} \approx m P(X, \Pi_X)$ and Assumption 2 through 4 hold, then $g(X, \phi) > g(X, \{Y^*\})$.*

Proposition 5 allows us to conclude that K2 will not be able to learn a PI model correctly.

## 5 REMARKS

We have shown that for a probabilistic domain of any size, there exists a collection of PI models where local independence does not imply global independence. Consequently, single link lookahead search that relies on local dependence to identify collective dependency will fail to learn PI models. We have demonstrated this failure in a number of learning algorithms which employ different scoring metrics. When such an incorrectly learned model is used in the inference, the posteriors obtained will be incorrect whenever the set of query and evidence variables covers a PI submodel.

The existence of PI models poses a challenge to learning probabilistic networks as approximate I-maps. Our result suggests that when prior knowledge about a domain does not eliminate the possible existence of a PI model, a multi-link lookahead search or other heuristics should be used. It appears that when single link lookahead search is used to learn a PI model, the learned network will be disconnected. This provides a warning for the possibility of an underlying PI model but may also signify a PDM with truly independent components. On the other hand, if the learned network is connected, it may be sufficient to eliminate the possibility of a PI model. Some proposals for managing the learning of PI models is given in [Xiang 95] and further research is needed.

In our discussion on PI models (Theorem 1), a special type of PI model (binary and 0.5 marginals) was used. A general characterization of all PI models is being attempted, which in turn will shed insight into the design of better learning algorithms.

At the moment, we do not have sufficient statistics on the frequency of PI model in practical problem domains. It is worthwhile to search through data from different problem domains for an empirical indication of such frequency. However, *parity* functions are well known to cause failure of many decision tree learning algorithms, see for example [Pagallo 90]. PI models are a generalization of parity functions. Table 1 is a parity function, but Tables 2, 3 and 4 are *not*.

## Acknowledgement

This work is supported by Research Grant OGP0155425 from the Natural Sciences and Engineering Research Council (NSERC). The authors are members of the Institute for Robotics and Intelligent Systems (IRIS) and wish to acknowledge the support of the Networks of Centres of Excellence Program of the Government of Canada, NSERC, and the participation of PRECARN Associates Inc.